\documentclass[letterpaper, 10pt, conference]{ieeeconf}      

\IEEEoverridecommandlockouts                              
\usepackage{amsmath} 
\usepackage{amssymb}  
\usepackage{algorithm}
\usepackage{algorithmic}
\usepackage{graphicx}
\graphicspath{{}}
\usepackage{epstopdf}

\newtheorem{thm}{Theorem}[section]
\newtheorem{cor}[thm]{Corollary}
\newtheorem{lem}[thm]{Lemma}
\newtheorem{remark}[thm]{Remark}

\usepackage[{hmargin={54pt, 54pt}, vmargin={54pt,55pt}}]{geometry}

\title{\LARGE \bf
Parametrized Stochastic Multi-armed Bandits with Binary Rewards
}

\author{Chong Jiang and R. Srikant\\
Coordinated Science Laboratory and\\
Dept of Electrical and Computer Engineering\\
University of Illinois at Urbana-Champaign\\
Email: \{jiang17,rsrikant\}@illinois.edu%
\thanks{Research supported in part by AFOSR MURI FA 9550-­10-­1­-0573.}%
}

\begin{document}
\maketitle
\thispagestyle{empty}
\pagestyle{empty}

\begin{abstract}
In this paper, we consider the problem of multi-armed bandits with a 
large, possibly infinite number of correlated arms. We assume that the arms have Bernoulli distributed rewards, independent 
across time, where the probabilities of success are parametrized by known 
attribute vectors for each arm, as well as an unknown preference vector, each of dimension $n$. 
For this model, we seek an algorithm with a total regret that 
is sub-linear in time and independent of the number of arms. We present 
such an algorithm, which we call the Two-Phase Algorithm, and analyze 
its performance. We show upper bounds on the total regret which applies uniformly in time, for both the finite and infinite arm cases.
The asymptotics of the finite arm bound show that for any $f \in \omega(\log(T))$, the total 
regret can be made to be $O(n \cdot f(T))$. In the infinite arm case, the total regret is $O( \sqrt{n^3 T} )$.

\end{abstract}
\section{INTRODUCTION}
\subsection{Motivation}
The stochastic multi-armed bandit problem is the following: suppose
we are allowed to choose to ``pull,'' or play, any one of $m$
slot machines (also known as one-armed bandits) in each of $T$ timesteps,
where each slot machine generates a reward according to its own distribution
which is unknown to us. The parameters of the reward distributions are correlated between machines, 
but the rewards themselves are independent across machines, 
and independent and identically distributed across timesteps. The choice of which arm to pull 
may be a function of the sequence of past pulls and the sequence of
past rewards. If our goal is to maximize the total reward obtained,
taking expectation over the randomness of the outcomes, ideally we would
pull the arm with the largest mean at every timestep. However, we do not
know in advance which arm has the largest mean, so a certain amount
of exploration is required. Too much exploration, though, wastes time that could be spent 
reaping the reward offered by the best arm. This exemplifies
the fundamental trade-off between exploration and exploitation present
in a wide class of online machine learning problems. 

We consider a model for multi-armed bandit problems in which a
large number of arms are present, where the expected rewards of the
arms are coupled through an unknown parameter of lower dimension.
Now, it is no longer necessary for each arm to be investigated
in order to estimate the expected reward from that arm. Instead, we
can estimate the underlying parameter; in this way, each pull can yield
information about multiple arms. We present a simple algorithm, 
as well as bounds on the expected total regret as a function of 
time horizon when using this algorithm. While possibly sub-optimal, 
these bounds are independent of the number of arms.

This model is applicable to certain e-commerce applications: 
suppose an online retailer has a large number of related products,
and wishes to maximize revenue or profit coming from a certain set
of customers. If the preferences of this set of customers are known,
the list of items which are displayed can be sorted in descending
order of expected revenue or profit. However, we may not know a priori
what this preference vector is, so we wish to learn online 
by sequentially presenting each user with an item, observing whether
the user buys the item, and then updating an internal estimate of
the preference vector. 


As a concrete example, imagine an online camera store, with hundreds
of different camera models in stock. However, there are perhaps closer
to ten features which people will compare when deciding which, if
any, to purchase. There are permanent features of the camera itself,
such as megapixel count, brand name, and year of introduction, as
well as extrinsic features, such as price, review scores, and item
popularity. All of these features might be considered by the customer in 
order to decide whether or not to buy the camera. If bought, the store gains
a profit corresponding to the item. A key distinction of our model,
when compared to previous work, is the incorporation of this inherently
binary choice customers are faced with: to buy or not to buy.


\subsection{Model}

Our model consists of a multi-armed bandit with a set $U$ consisting of $m$ arms (items)
and $n$ underlying parameters (attributes), where $m \ge n$ and potentially
$m \gg n$. We will interchangeably also think of $U$ as being a $n \times m$ matrix, where each arm $u$ is an $n$-dimensional
attribute vector, and is one of the columns of $U$. Furthermore, we will assume that $\mbox{rank}(U)=n$.
There is also a constant but unknown preference vector
$z^{*}\in\mathbb{R}^{n}$. The quality $\beta_{u}=u^{T}z^{*}$
of arm $u$ is a scalar indicating how desirable the item is to a user. 
We will use the logistic function $f$ to define the expected
reward of an arm $u$, assuming a particular $z$, as 
$$\alpha_{u}(z)=f \left(u^{T}z\right) = \dfrac{1}{1+\exp\left(-u^{T}z\right)}.$$
Thus, the expected rewards of all of the arms are coupled through $z^*$.
For notational simplicity, we define $\left. \alpha_{u}^{*}=\alpha_{u}(z^{*}) \right.$.
Let the set of equally best arms be
$$ V = \left\{ v \in U : \alpha_v^* = \max_{u \in U} \alpha_u^* \right\} \subset U .$$
Define the expected reward of a best arm to be
$$ \alpha_V^* = \max_{u \in U} \alpha_u^* .$$
At each timestep $t$ up to a finite time horizon $T$,
a policy will choose to pull exactly one arm, call this arm $C_{t}$,
and a reward $X_{t}$ will be obtained, where $X_{t}\sim\mbox{Ber}(\alpha_{C_{t}}^{*})$.
We wish to find policies $g$ which maximize the total expected reward, 
$\sum_{t=1}^{T}X_{t}$, or equivalently, minimize the expected total 
regret, $E_{g}\left[\sum_{t=1}^{T} \left( \alpha_V^* - X_{t} \right) \right]=T\cdot\alpha_{V}^{*}-E_{g}[\sum_{t=1}^{T}\alpha_{C_{t}}^{*}]$.
\subsection{Prior Work}
For an introduction and survey of classical multi-armed bandit problems and their variations, see Mahajan and Teneketzis \cite{mahajan_teneketzis_07}. One of the earliest breakthroughs on the classical multi-armed bandit problem came from Gittins and Jones \cite{gittins_jones_74}, who showed that under geometric discounting, the optimal policy assigns an index to each arm, now known as the Gittins index, and pulls the arm with the largest Gittins index. Other proofs of this optimality have been given later by Weber \cite{weber_92} and Tsitsiklis \cite{tsitsiklis_94}. Whittle \cite{whittle_88} proved that a similar index-based result is nearly optimal in the ``restless bandit'' variation of this model, where the arms which are not pulled also evolve in time. While these policies greatly simplify a single $m$-dimensional problem into $m$ 1-dimensional problems, it is still, in general, too computationally complex for online learning.

Lai and Robbins \cite{lai_robbins_85} proved an achievable $O(m \cdot \log T)$
lower bound for the expected total regret of the stochastic multi-armed bandit problem in the case of independent arms. Related work by Agrawal \textsl{et al.} \cite{agrawal_teneketzis_anantharam_89_iid, agrawal_teneketzis_anantharam_89_markov,agrawal_hegde_teneketzis_87,agrawal_95} and Anantharam \textsl{et al.} \cite{anantharam_varaiya_walrand_87_iid, anantharam_varaiya_walrand_87_markov} considered similar models with i.i.d. and Markov time dependencies for each arm, constructed index policies which are computationally much simpler, and extended the results to include ``multiple plays'' and ``switching costs''.

Abe \textsl{et al.} \cite{abe_biermann_long_03} and Auer \cite{auer_03} considered models with finite numbers of arms, with reward distributions that are correlated through a multi-variate parameter $z$ of dimension $n$, and obtained upper bounds on the regret of order $O(\sqrt{m T})$ and $O(\sqrt{n T}\cdot \log T)$, respectively. 
Mersereau \textsl{et al.} \cite{merserau_rusmevichientong_tsitsiklis_08} considered a model
in which the expected rewards are affine functions of a scalar parameter $z$, but allowed the set of arms to be a bounded, convex region in $\mathbb{R}^{n}$, in which case $m$ is uncountably infinite.
They then derived a policy whose expected total regret is $\Theta(\sqrt{T})$. Rusmevichientong and Tsitsiklis \cite{rusmevichientong_tsitsiklis_08} expanded this model to allow for a multi-variate parameter $z$ 
of dimension $n$, and showed that the expected total regret (ignoring
$\log T$ factors) is $\Theta(n\sqrt{T})$. Dani \textsl{et al.} \cite{dani_hayes_kakade_08} independently considered a nearly identical model, and obtained similar results. Kleinberg \textsl{et al.} \cite{kleinberg_slivkins_upfal_metricspaces_08} considered a model in which the deterministic rewards are a Lipschitz-continuous function of the $n$-dimensional vector corresponding to each arm, and obtain an expected total regret (ignoring
$\log T$ factors) of $\Theta(T^{\frac{n+1}{n+2}})$.

Auer \textsl{et al.} \cite{auer_cesabianchi_freund_schapire_02} considered
a non-stochastic version of the multi-armed bandit problem, in which
the rewards are no longer drawn from an unknown distribution, but
can instead be adversarially generated. The resultant total weak regret,
calculated by comparison with the single arm which is best
over the entire time horizon, is shown to be $O(\sqrt{mT})$.
The change from logarithmic to polynomial regret in this model is
due to having rewards which are time-dependent and potentially adversarially 
generated, instead of being drawn from a time-independent distribution.

Audibert \textsl{et al.} \cite{audibert_bubeck_munos_10} considered the problem of best arm identification in a 
stochastic multi-armed bandit setting, but where the goal is to maximize the probability of 
determining the best arm at the end of a time horizon, as opposed to the usual goal of 
minimizing total regret over a time horizon. This model is useful when considering 
exploration and exploitation as occurring in series, instead of in parallel. The probability of error is shown to be upper bounded by a decaying exponential in $T$.

Auer \textsl{et al.} \cite{auer_cesabianchi_fischer_02} investigated the finite-time regret of the multi-armed bandit problem, assuming bounded but otherwise arbitrary reward distributions. Using
upper confidence bound (UCB) algorithms, where the confidence interval of an arm shrinks as the arm is subjected to more plays, they achieve a logarithmic upper bound on the regret, uniform over time, that scales with the ``gaps'' between the expected rewards for the arms. One algorithm they propose, UCB2, selects the arm with largest empirical mean plus confidence interval, plays it for a number of timesteps dependent on how often that particular arm has been selected in the past, and repeats this process until the time-horizon is reached. This achieves asymptotically optimal expected total regret, and has the best constant possible.

A common idea used in crafting policies to solve the multi-armed bandit problem is that of the doubling trick \cite{stoltz, prediction_learning_games}. This technique is used to 
convert a parametrized algorithm which works on a time horizon $T$, along with its corresponding bound, into a non-parametrized algorithm that runs forever, with an upper bound that holds uniformly over time.
\section{TWO-PHASE ALGORITHM}

We first present an algorithmic description of a policy for the multi-armed bandit problem described in Section I.B. This algorithm, which we call the Two-Phase Algorithm, will depend on a scheduling function $\left. g:\mathbb{N}_1 \rightarrow \mathbb{N}_0 \right.$, such that $g$ is strictly increasing.
Since $g$ is not surjective in general, its inverse $g^{-1}$ is not defined over all of $\mathbb{N}_0$; however, we can extend the inverse image in the natural way to preserve monotonicity, by defining
$\left. g^{-1}: \mathbb{N}_0 \rightarrow \mathbb{N}_1 \right.$,
$$g^{-1}(t) = 
\max \left\{ 1 \cup \left\{ l \in \mathbb{N}_1 : g(l) \le t \right\} \right\} .
$$
In Theorem \ref{theorem finite regret}, we will show an upper-bound to the expected total regret of this policy on finite arms, which is independent of the number of arms $m$.
In Theorem \ref{theorem infinite regret}, we will show an upper-bound to the expected total regret of this policy on a special case when there are uncountably infinite arms.

\begin{algorithm}
\caption{Two-Phase Algorithm}
\label{two_phase}
\begin{algorithmic}[1]
\REQUIRE Set of all arms $U$
\REQUIRE Set of $n$ chosen arms $\Sigma = \left\{ \Sigma_1, \ldots, \Sigma_n \right\} \subseteq U$, s.t. $\Sigma$ has rank $n$
\REQUIRE Scheduling function $g : \mathbb{N}_1 \rightarrow \mathbb{N}_0$, strictly increasing
\STATE $t \leftarrow 1$, $l \leftarrow 1$
\STATE $q_{u} \leftarrow 0, \forall u \in \Sigma$
\LOOP
	\FOR{$u \in \Sigma$}
		\STATE Pull arm $C_t \leftarrow u$, obtain reward $X_t$ \COMMENT{Phase 1}
		\STATE $q_{C_t} \leftarrow q_{C_t} + 1_{\{X_t\}}$
		\STATE $t \leftarrow t+1$
	\ENDFOR
		
	\STATE Form the estimates $\hat{\alpha}_{u,l} \leftarrow \dfrac{q_{u}}{l},\ \forall u\in\Sigma$
	\IF{ $\hat{\alpha}_{u,l} \in (0,1), \ \forall u \in \Sigma$ }
		\STATE $\hat{z}_l \leftarrow \left(\Sigma^{T}\right)^{-1}\left[\begin{array}{c}
														f^{-1}\left(\hat{\alpha}_{\Sigma_1,l}\right)\\
														\vdots\\
														f^{-1}\left(\hat{\alpha}_{\Sigma_n,l}\right)\end{array}\right]$
	\ELSE 
		\STATE $\hat{z}_l \leftarrow \boldsymbol{0}_n$
	\ENDIF
	
	\STATE $C_{(l)} \leftarrow \arg\max_{u \in U} \alpha_u (\hat{z}_l)$, settling ties arbitrarily
		
	\FOR{$s \leftarrow 1$ \TO $g(l)$}
		\STATE Pull arm $C_t \leftarrow C_{(l)}$, obtain reward $X_t$ \COMMENT{Phase 2}
		\STATE $t \leftarrow t+1$
	\ENDFOR
	\STATE $l \leftarrow l+1$
\ENDLOOP
\end{algorithmic}
\end{algorithm}
\begin{figure*}
\begin{center}
\includegraphics{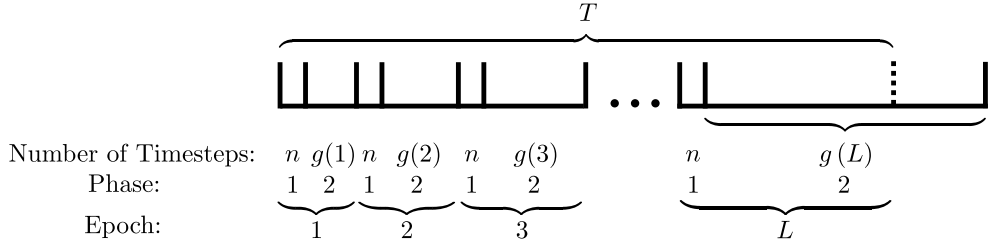}
\end{center}
\caption{Given a time horizon $T$, we partition the $T$ timesteps into Phase 1 and Phase 2 timesteps, grouped into a total of $L$ epochs.}
\label{fig:two_phase_timeline}
\end{figure*}
The algorithm requires a selection of $n$ arms,
$$\Sigma = \left\{ \Sigma_1, \ldots, \Sigma_n \right\} \subseteq U, \mbox{ s.t. } \Sigma \mbox{ has rank } n .$$
Such a choice exists since we assume $U$ has rank $n$.
The algorithm proceeds in epochs; epoch $l$ consists of $n$ exploration pulls (called Phase 1), one for each arm in $\Sigma$, and $g(l)$ exploitation pulls (called Phase 2). In other words, Phase 1 refines our estimate of $z^*$, and Phase 2 repeatedly pulls the best arm given our current estimate $\hat{z}_l$. If we impose a time horizon of $T$, epochs ${1, 2, \ldots, L}$ are appended until the time horizon $T$ has been reached. The two phases are illustrated in Figure \ref{fig:two_phase_timeline}.

For each timestep $t$ in Phase 1, an arm $u \in \Sigma$ is chosen, and the empirical count of successes $q_{u}$ is incremented if $X_t = 1$.
Prior to each Phase 2 timestep during epoch $l$, there have already been $l$ Phase 1 pulls. We
can then form empirical estimates for $\alpha_{i}^{*}$ based on the Phase 1 timesteps, namely  $\hat{\alpha}_{u,l}=\dfrac{q_{u}}{l},\ \forall u\in\Sigma$. If 
$$\hat{\alpha}_{u,l} \in (0,1), \ \forall u \in \Sigma ,$$
then we call epoch $l$ a good epoch, and form the current best estimate for $z^*$,
$$
\hat{z}_l = \left(\Sigma^{T}\right)^{-1}\left[\begin{array}{c}
														f^{-1}\left(\hat{\alpha}_{\Sigma_1,l}\right)\\
														\vdots\\
														f^{-1}\left(\hat{\alpha}_{\Sigma_n,l}\right)\end{array}\right],
$$
since $f$ being strictly increasing and continuous implies $f^{-1}$ exists on $(0,1)$, and since $\Sigma$ being an $n \times n$ matrix with full rank implies $\left(\Sigma^{T}\right)^{-1}$ exists. Otherwise, we call epoch $l$ a bad epoch, and let $\hat{z}_l = \boldsymbol{0}_n$. Define the event $G_l$ to mean that epoch $l$ is a good epoch. Note that $G_l \implies G_{l+i}\ \forall i \in \mathbb{N}_1$.

Then, choose an arm
$$C_{(l)}=\arg\max_{u \in U} \alpha_{u}(\hat{z}_l) ,$$
settling ties arbitrarily, and pull this arm $g(l)$ times to form the current epoch's Phase 2.

\begin{remark}
In practice, LU decomposition, instead of matrix inversion, can be used to solve for $\hat{z}_l$. Also, since $f$ is strictly increasing, the estimated best arm in a good epoch $l$ can be computed as 
$$C_{(l)} = \arg\max_{u \in U}\left(u^T \hat{z}_l \right) .$$
\end{remark}

We shall point out some of the ideas behind this algorithm. First, the algorithm is defined to run indefinitely; to obtain the total regret for any finite time horizon $T$, we simply terminate the algorithm when timestep $T$ has been reached. This achieves the same outcome as an application of the doubling trick, in that the algorithm is not dependent on a time horizon $T$. Our algorithm is similar to the algorithm UCB2 of \cite{auer_cesabianchi_fischer_02}. The main difference is that in our exploration phases, the choice of arm exploits the correlation model that we have assumed in our problem. Furthermore, as we will see later, unlike UCB2, the lengths of the exploitation phases are chosen to grow sub-exponentially in the epoch number (e.g., $g(l) \in \exp \left( o(l) \right)$ for finite arms) in order to obtain a regret bound that grows slightly faster than logarithmically in the time horizon (e.g., $ E[R_T] \in \omega ( \log(T))$ for finite arms). As we gain more information and are able to estimate $z^*$ more accurately, we can spend a greater fraction of timesteps exploiting the arm we think is best; this is achieved by choosing a suitable scheduling function $g$ to control the ratio of the number of exploitation (Phase 2) pulls versus exploration (Phase 1) pulls, as a function of the epoch number $l$.

Note that there is only randomness in the outcomes $\left\{ X_t \right\}_{t=1}^{T}$, since the Two-Phase Algorithm is deterministic in the selection of the arm $C_t$, conditioned on the history. We will use $\omega$ to denote the sample-paths of $\left\{ X_t \right\}_{t=1}^{T}$. Let $L$ denote the number of epochs (including partial epochs, as the final one may be truncated) up to timestep $T$, which is independent of sample-path $\omega$. While $L$ is actually a function of $T$, we will not write this dependence explicitly.

Define the expected regret in a single Phase 2 timestep in epoch $l$ to be $E[r_{2,l}]$. Note that this value is the same for every Phase 2 timestep in epoch $l$, and hence is independent of timestep.
Define the total regret up to timestep $T$ in the Phase $i$ timesteps for a sample-path $\omega$ to be 
$R_{i,T}\left(\omega\right) .$
Define the total regret up to timestep $T$ for a sample-path $\omega$ to be
$\left. R_{T}\left(\omega\right)=R_{1,T}\left(\omega\right)+R_{2,T}\left(\omega\right) . \right.$
Our goal is to find an upper-bound on $E\left[R_{T}\right]$, the expected total regret. In particular, we are interested in the asymptotic behavior of the upper-bound as $T \rightarrow \infty$.

\section{ANALYSIS, FINITE ARMS}

Consider the multi-armed bandit problem described in Section I.B, where the $m$ arms can each take on any value in $\mathbb{R}^n$, as long as $U$ is full rank.

\subsection{Upper-bound Results}

\begin{lem}\label{lemma total phase 1}
For the Two-Phase Algorithm, we have the following bound on the expected total Phase 1 regret up to timestep $T$: 
$$
E\left[R_{1,T} \right] \le \alpha_V^{*} n L.
$$
\end{lem}

Proof:
\begin{align*}
E \left[ R_{1,T} \right] & \le E \left[ \sum_{l=1}^{L} \sum_{u \in \Sigma} \left( \alpha_V^{*} - \alpha_{u}^{*} \right) \right] \\
 & \le \alpha_V^{*} n L.
\end{align*}
\begin{flushright}
$\square$
\end{flushright}

\begin{figure}
\begin{center}
\includegraphics{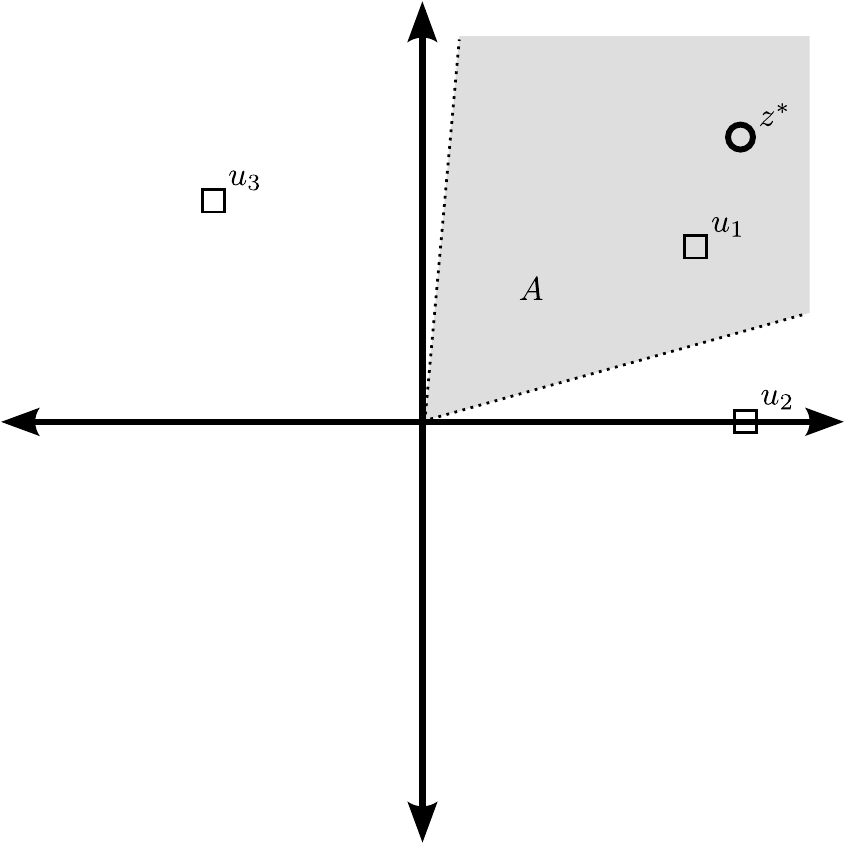}
\end{center}
\caption{As an example, consider a scenario with $n=2$ and $m=3$. The arms $U = \left\{ u_1, u_2, u_3 \right\}$ and the preference vector $z^{*}$ are located at the indicated points. 
The shaded region is $A$; the boundary of $A$ is formed by the perpindicular bisectors of the segments $u_3, u_1$ and $u_1, u_2$.}
\label{fig:three_phase_A}
\end{figure}

\begin{lem}\label{lemma bad epoch}
The probability that epoch $l$ is a bad epoch is upper-bounded by 
$$2n \cdot \exp\left\{ -2l\cdot  f \left( - \left\Vert z^{*}\right\Vert \right) \right\}.$$
\end{lem}

Proof:
In order for epoch $l$ to be a good epoch, we have the condition that $\hat{\alpha}_{u,l} \in (0, 1) \ \forall u \in \Sigma$.
Consider the condition for a bad epoch:
\begin{align*}
& \exists u \in \Sigma \mbox{ s.t. } \hat{\alpha}_{u,l} \notin (0, 1) \\
& \implies \exists u \in \Sigma \mbox{ s.t.} \left| \hat{\alpha}_{u,l} - \frac{1}{2} \right| \ge \frac{1}{2} \\
 & \implies \exists u \in \Sigma \mbox{ s.t.} \left| \hat{\alpha}_{u,l} - \alpha_u^* \right| \ge \frac{1}{2} - \left| \alpha_u^* - \frac{1}{2} \right| \\
 & \implies \exists u \in \Sigma \mbox{ s.t.} \left| \hat{\alpha}_{u,l} - \alpha_u^* \right| \ge \frac{1}{2} - 
 \max_{v \in \Sigma}\left| \alpha_v^* - \frac{1}{2} \right|.
\end{align*}
Note that
$$\max_{v \in \Sigma}\left| \alpha_v^* - \frac{1}{2} \right| \le f \left( \max_{u \in \Sigma} \left\Vert u \right\Vert \cdot \left\Vert z^{*}\right\Vert \right) - \frac{1}{2} .$$
Then, applying the union bound and Chernoff bound, it follows that
\begin{align*}
& P \left( \exists i \in \Sigma \mbox{ s.t. } \hat{\alpha}_{i,l} \notin (0, 1) \right) \\
\le & 2n \cdot \exp\left\{ -2l\cdot \left[ 1 - f \left( \max_{u \in \Sigma} \left\Vert u \right\Vert \cdot \left\Vert z^{*}\right\Vert \right) \right] 
\right\} \\
= & 2n \cdot \exp\left\{ -2l\cdot  f \left( - \max_{u \in \Sigma} \left\Vert u \right\Vert \cdot \left\Vert z^{*}\right\Vert \right) \right\}.
\end{align*}
Let $k_1 = 2 f \left( \max_{u \in \Sigma} \left\Vert u \right\Vert \cdot \left\Vert z^{*}\right\Vert \right)$.
Thus, the probability that epoch $l$ is a bad epoch is then upper-bounded by 
$$2n \cdot \exp ( -k_1 l ) .$$
\begin{flushright}
$\square$
\end{flushright}

\begin{lem}\label{lemma finite per timestep phase 2}
For the Two-Phase Algorithm on finite arms, for a given choice of scheduling function 
$$g \mbox{ s.t. } g(l) \in \exp \left( o(l) \right) ,$$ 
we have the following bound on the expected Phase 2 regret per timestep in a good epoch $l$:
$$
E \left[r_{2,l} | G_l \right] \le 2\alpha_V^* n \cdot \exp\left(- \gamma l \right) ,
$$
where $\gamma$ is a constant which depends on $U$ and $z^*$.
\end{lem}

Proof:
Recall that $\alpha_u^* = f \left( u^T z^* \right)$, where
$$f \left( \beta \right) = \dfrac{1}{1+\exp\left(-\beta\right)}$$
is strictly increasing and continuous. Thus $f^{-1}$ is well defined, strictly increasing and continuous.
Recall that
$$ V = \left\{ v \in U : \alpha_v^* = \max_{u \in U} \alpha_u^* \right\} $$
is the set of equally best arms. Because $f \left( u_i^T z \right)$ is continuous in $z$ and defined over $\mathbb{R}^n$, it follows that there exists a neighborhood of $z^*$, denoted $A$, such that 
$$A = \left\{ z \in \mathbb{R}^n : \arg\max_{u \in U} \alpha_u(z) \in V \right\}.$$
Since $\Sigma$ is full rank, $A$ must contain an open parallelotope centered at $z^*$, 
$$B_{z^*} \left( \delta \right) = \left\{z \in \mathbb{R}^n : \left\Vert \Sigma^T z - \Sigma^T z^* \right\Vert _{\infty} < \delta \right\},$$
where $\delta > 0$ and is largest possible. An example of the problem parameters and the induced region $A$ is shown in Figure \ref{fig:three_phase_A}.

Consider any $z \in B_{z^*} \left( \delta \right)$. By definition, 
$$|u^T z - u^T z^*| < \delta,\  \forall u \in \Sigma.$$
This is equivalent to
$$\left| f^{-1} \left( \alpha_u(z) \right) - f^{-1} \left( \alpha_u^* \right) \right| < \delta,\ \forall u \in \Sigma.$$
Since $f^{-1}$ is continuous, this is equivalent to having a set of constants 
\begin{align*}
\left\{ \underline{\alpha}_u, \overline{\alpha}_u \right\} _{u \in \Sigma} \mbox{ s.t. }
\underline{\alpha}_u < \alpha_u(z) < \overline{\alpha}_u ,
\end{align*}
where 
\begin{align*}
& \underline{\alpha}_u = f \left( f^{-1} \left( \alpha_u^* \right) - \delta \right) \mbox{ and} \\
& \overline{\alpha}_u = f \left( f^{-1} \left( \alpha_u^* \right) + \delta \right) ,\ \forall u \in \Sigma .
\end{align*}
For a Phase 2 timestep during a good epoch $l$, the algorithm forms the empirical average rewards 
$$\hat{\alpha}_{u,l} \in (0, 1),\ \forall u \in \Sigma .$$ 
By the discussion above,
\begin{align*}
& \underline{\alpha}_u < \hat{\alpha}_{u,l} < \overline{\alpha}_u, \ \forall u \in \Sigma \\
\implies & \left. \hat{z}_l \in B_{z^*} (\delta) \subseteq A \right. \\
\implies & \left. C_t = \arg\max_{u \in U} \{u^T \hat{z}_l\} \in V \right.
\end{align*}
and we will have chosen one of the best arms, accumulating zero regret.

Note that during epoch $l$, $\hat{\alpha}_{u,l}$
is a sum of $l$ i.i.d. $\mbox{Ber}\left(\alpha_{u}^{*}\right)$
random variables, $\forall u\in\Sigma$. By the Chernoff bound,
\begin{align*}
P\left(\hat{\alpha}_{u,l} < \underline{\alpha}_{u} | G_l \right) 
\le & \exp\left[-l \cdot D\left(\underline{\alpha}_{u}||\alpha_{u}^{*}\right)\right] \mbox{, and} \\
P\left(\hat{\alpha}_{u,l} > \overline{\alpha}_{u} | G_l \right) 
\le & \exp\left[-l \cdot D\left(\overline{\alpha}_{u}||\alpha_{u}^{*}\right)\right],\ \forall u \in \Sigma,
\end{align*}
where $D\left(p||q\right)= p \cdot \log \dfrac{p}{q} + \left( 1-p \right) \cdot \log \dfrac{1-p}{1-q}$ is the K-L divergence between two Bernoulli
distributions.

Let $\gamma=\min _{u \in \Sigma} \min \left\{ D\left(\underline{\alpha}_{u}||\alpha_{u}^{*}\right) , D\left(\overline{\alpha}_{u}||\alpha_{u}^{*}\right)\right\}$.
Note that from the definitions of $\underline{\alpha}_{i}$ and $\overline{\alpha}_{i}$, it follows that 
$$\underline{\alpha}_u < \alpha_u^* < \overline{\alpha}_u, \ \forall u
 \in \Sigma.$$
Since $D\left(p||q\right)=0\iff p=q$, we have that
$\gamma>0$. 
By the union bound, 
$$P\left(\exists u\in\Sigma:\hat{\alpha}_{u,l}\notin\left(\underline{\alpha}_{u},\overline{\alpha}_{u}\right) | G_l \right) \le 2n \cdot \exp\left(-\gamma l\right).$$

Reviewing the chain of implications, we have 
\begin{align*}
& P\left(\hat{z}_l \notin A | G_l \right) \\
\le & P\left(\hat{z}_l \notin B_{z^*}(\delta) | G_l \right) \\
 = & P\left( \left\Vert \Sigma^{T} \hat{z}_{l} - \Sigma^{T} z^{*} \right\Vert_{\infty} > \delta | G_l \right) \\
 = & P\left( \exists u \in \Sigma: \left| u^{T}\hat{z}_l - u^{T}z^{*} \right| > \delta | G_l \right) \\
 = & P\left( \exists u \in \Sigma:\left| f^{-1}\left(\hat{\alpha}_{u,l}\right) - f^{-1}\left(\alpha_u^{*}\right) \right|>\delta | G_l \right)\\
 = & P\left( \exists u \in \Sigma:\hat{\alpha}_{u,l} \notin \left( \underline{\alpha}_u ,\overline{\alpha}_u \right) | G_l \right)\\
 \le & 2n \cdot \exp\left(- \gamma l \right).
 \end{align*}

Then, we have a bound on the expected per-timestep regret $r_{2,l}$ during Phase 2 of epoch $l$:
\begin{align*}
E \left[ r_{2,l} | G_l \right] = & E\left[r_{2,l} | \hat{z}_l \in A, G_l \right] \cdot P\left(\hat{z}_l \in A | G_l \right) \\
 & + E\left[r_{2,l} | \hat{z}_l \notin A | G_l \right] \cdot P\left(\hat{z}_l \notin A | G_l \right)\\
 \le & 0\cdot P\left(\hat{z}_l \in A | G_l \right) + \alpha_V^* \cdot P\left(\hat{z}_l \notin A | G_l\right)\\
 \le & 2\alpha_V^* n \cdot \exp\left(-\gamma l \right) .
\end{align*}
\begin{flushright}
$\square$
\end{flushright}

\begin{lem}\label{lemma finite total phase 2}
For the Two-Phase Algorithm on finite arms, for a given choice of scheduling function 
$$g \mbox{ s.t. } g(l) \in \exp \left( o(l) \right) ,$$ 
we have the following bound on the expected total Phase 2 regret up to timestep $T$:
$$
E[R_{2,T}] \le \alpha_V^* n \left( 2 k_2 + L \right) ,
$$
where $k_2$ is a constant which depends on $U$ and $z^{*}$.
\end{lem}

Proof:
By Lemmas \ref{lemma bad epoch} and \ref{lemma finite per timestep phase 2}, we can upper-bound the expected total Phase 2 regret,
\begin{align*}
& E[R_{2,T}] \\
\le & \sum_{l=1}^{L} \left\{ \Big[ P(G_l) \cdot E[r_{2,l}|G_l] + P( \neg G_l) \cdot E[r_{2,l}| \neg G_l] \Big] \cdot g(l) \right\} \\
\le & \sum_{l=1}^{L} \left\{ \left[ P(G_l) \cdot 2\alpha_V^* n \cdot \exp\left(-\gamma l \right) + P( \neg G_l) \cdot \alpha_V^* \right] \cdot g(l) \right\} \\
\le & \sum_{l=1}^{L} \left\{ \left[ 2\alpha_V^* n \cdot \exp\left(- \gamma l \right) + 2 \alpha_V^* n \cdot \exp\left( -k_1 l \right) \right] \cdot g(l) \right\} \\
\le & 2\alpha_V^* n \sum_{l=1}^{L'} \left\{ \left[ \exp\left(- \gamma l \right) +\exp\left( -k_1 l \right) \right] \cdot g(l) \right\} \\
& + 2\alpha_V^* n \sum_{l=L'+1}^{L} \dfrac{1}{2} \\
\le & \alpha_V^* n \left\{ 2\sum_{l=1}^{L'} \left\{ \left[ \exp\left(- \gamma l \right) +\exp\left( -k_1 l \right) \right] \cdot g(l) \right\} + L \right\} ,
\end{align*}
where 
$$L'=\max\left\{ l: \left[ \exp\left(- \gamma l \right) +\exp\left( -k_1 l \right) \right] \cdot g(l) > \dfrac{1}{2}\right\}$$ 
is a constant, independent of sample-path, that depends on $U$ and $z^{*}$ (and
is therefore unknown to the algorithm). However, since we have assumed $g(l) \in \exp \left( o(l) \right)$,
it follows that 
$$\lim_{l \rightarrow \infty} \left[ \exp\left(- \gamma l \right) +\exp\left( -k_1 l \right) \right] \cdot g(l) = 0 ,$$
and thus $L'$ is finite. 
Let 
$$k_2 = \sum_{l=1}^{L'} \left\{ \left[ \exp\left(- \gamma l \right) +\exp\left( -k_1 l \right) \right] \cdot g(l) \right\}, $$
which is well defined since $L'$ is finite.
Thus,
$$ E[R_{2,T}] \le \alpha_V^* n \left( 2 k_2 + L \right) . $$
\begin{flushright}
$\square$
\end{flushright}

\begin{thm}\label{theorem finite regret}
For the Two-Phase Algorithm on finite arms, for a given choice of scheduling function 
$$g \mbox{ s.t. } g(l) \in \exp \left( o(l) \right) ,$$ 
we have the following bound on the expected total regret up to time-horizon $T$:
$$ E \left[ R_{T} \right] \le 2 \alpha_V^* n \left( k_2 +  g^{-1}(T) + 1 \right) . $$
\end{thm}

Proof:
Since the final epoch may be only partially finished, we will lower-bound the total time with the number of timesteps in the penultimate epoch's Phase 2,
\begin{align*}
T & \ge \sum_{l=1}^{L-1} \left\{ n + g(l) \right\} \ge g(L-1).
\end{align*}
Equivalently,
$$L \le g^{-1}(T) + 1.$$
Then, using Lemmas \ref{lemma total phase 1} and \ref{lemma finite total phase 2},
\begin{align*}
E[R_{T}] = & E[R_{1,T}] + E[R_{2,T}] \\
\le & 2 \alpha_V^* n \left( k_2 + L \right) \\
\le & 2 \alpha_V^* n \left( k_2 +  g^{-1}(T) + 1 \right) .
\end{align*}
\begin{flushright}
$\square$
\end{flushright}

\begin{cor}\label{corollary finite regret asymptotic}
For the Two-Phase Algorithm on finite arms, for a given choice of scheduling function 
$$g \mbox{ s.t. } g(l) \in \exp \left( o(l) \right) ,$$ we have the following asymptotic bound on the expected total regret up to time-horizon $T$:
$$E \left[ R_{T} \right] \in O\left( n \cdot g^{-1}(T) \right).$$
\end{cor}
Proof:
By Theorem \ref{theorem finite regret}, as a function of $T$,
\begin{align*}
E[R_T] \le & 2 \alpha_V^* n \left( k_2 +  g^{-1}(T) + 1 \right) \\
\in \  & O\left( n \cdot g^{-1}(T) \right),
\end{align*}
since $\alpha_V^* \le 1$, $k_2 > 0 $ is a constant dependent only upon $U$ and $z^*$, and $g^{-1}(T) \in \omega(1)$.
\begin{flushright}
$\square$
\end{flushright}

\begin{lem}\label{lemma g g inverse asymptotics}
$$g^{-1}(t)\in \omega \left( \log(t) \right) \implies g(l) \in \exp \left( o(l) \right).$$
\end{lem}
Proof: 
\begin{align*}
\lim_{t \rightarrow \infty} \dfrac{\log(t)}{ g^{-1}(t)}
& = \lim_{l\rightarrow\infty} \dfrac{\log ( g(l) )}{ g^{-1}(g(l))} \tag{1} \\
& = \lim_{l\rightarrow\infty} \dfrac{\log(g(l))}{ l} \tag{2} \\
& = 0 \tag{3} ,
\end{align*}
where (1) is by making the substitution $t = g(l)$, recalling that 
$g:\mathbb{N}_1 \rightarrow \mathbb{N}_0$ is strictly increasing by assumption, so
$\left. \lim_{l \rightarrow \infty} g(l) = \infty \right.$.
(2) is since by construction,
$$g^{-1}(g(l)) = l, \ \forall l \in \mathbb{N}_1. $$
Lastly, (3) is since $g^{-1}(t)\in \omega \left( \log(t) \right)$, so by definition,
$$\lim_{t \rightarrow \infty} \dfrac{g^{-1}(t)}{\log(t)} = \infty. $$
Hence  
$\log \left( g(l) \right) \in o(l)$, so $g(l) \in \exp \left( o(l) \right)$, and thus $g$ is a valid scheduling function.
\begin{flushright}
$\square$
\end{flushright}

Let $\log^{*}(x)$, the iterated logarithm function, be defined recursively by
$$
\log^{*}(x)=
\begin{cases}
0, & \mbox{if }x\le 1\\
1+\log^{*}\left(\log x\right), & \mbox{if } x>1
\end{cases}.$$

\begin{cor}\label{corollary gLLS}
The Two-Phase Algorithm can achieve
$\left. E[R_T] \in O \left( n \cdot \log(T) \cdot \log^{*} (T) \right) \right.$.
\end{cor}

Proof: Choose 
$$g_{LLS}(l) = \max \left\{ t \in \mathbb{N}_1 :  \log(t) \cdot \log^*(t) \le l \right\}.$$
Then,
$$g^{-1}_{LLS}(t) = \left\lfloor \log(t) \cdot \log^*(t) \right\rfloor,$$
$$\lim_{t \rightarrow \infty} \dfrac{g^{-1}_{LLS}(t)}{\log(t)} = \lim_{t \rightarrow \infty} \log^*(t) \rightarrow \infty .$$ Thus, $g_{LLS} \in \omega \left( \log(t) \right)$, and by Lemma \ref{lemma g g inverse asymptotics} and Corollary \ref{corollary finite regret asymptotic}, we have an achievable expected total regret of 
$$E[R_T] \in O \left( n \cdot g_{LLS}^{-1}(T) \right) \subseteq O \left( n \cdot \log(T) \cdot \log^{*} (T) \right).$$
\begin{flushright}
$\square$
\end{flushright}

\begin{remark}
\label{Conjecture on Near-optimality}
In accordance with other results, such as \cite{lai_robbins_85}, we suspect this problem has a lower bound that is asymptotically $c n \cdot \log(T)$, where $c$ is dependent on the problem parameters $U$ and $z^*$. If this is the case, then by including the term $\log^* (T)$, we are able to obtain an upper bound which is not tight, but within a factor of $\log^*(T)$, while avoiding a dependence on the problem parameters.
\end{remark}

\subsection{Generalization to Arm-dependent Rewards}

Suppose that each arm $u\in U$ has a potentially different value of the reward, so that instead of a $\{0,1\}$ reward, it has a $\{0,w_u\}$ reward. Furthermore, suppose that $\{w_u\}_{u \in U}$ is known. Several definitions must be generalized, namely in the model,
$$X_{t}\sim w_{C_{t}}\cdot\mbox{Ber}\left(\alpha_{C_{t}}^{*}\right) ,$$
$$ V = \left\{ v \in U : w_v \alpha_v^* = \max_{u \in U} w_u \alpha_u^* \right\} \subset U ,$$
$$ w_V \alpha_V^* = \max_{u \in U} w_u \alpha_u^* ,$$
$$A = \left\{ z \in \mathbb{R}^n : \arg\max_{u \in U} w_u \alpha_u(z) \in V \right\} ,$$
$$E[R_T] = E_{g}\left[\sum_{t=1}^{T} \left( w_u \alpha_V^* - X_{t} \right) \right] ,$$
and in the algorithm,
$$ C_{(l)} \leftarrow \arg\max_{u \in U} w_u \alpha_u (\hat{z}_l) .$$
Then, Theorem \ref{theorem finite regret} generalizes with only minor modifications to the proof, yielding
$$E[R_{T}] \le 2 w_V \alpha_V^* n \left( k_2 +  g^{-1}(T) + 1 \right) .$$
Corollaries \ref{corollary finite regret asymptotic} and \ref{corollary gLLS} also generalize, with the same results as before.

\section{ANALYSIS, INFINITE ARMS}

Consider the multi-armed bandit problem described in Section I.B, with each point on the unit sphere in $\mathbb{R}^n$ being an arm. Now, the number of arms is uncountably infinite, and the finite arm analysis from before no longer yields a useful bound; since there is no longer a gap between the best and second-best arms, the region $A$ degenerates into a line, causing $\gamma=0$ and $k_2 = \infty$.

\subsection{Upper-bound Results}
For this special case of this infinite arms problem, we shall show that a total expected regret, up to time $T$, of
$O\left( \sqrt{n^3 T} \right)$ is achievable.
To obtain a meaningful bound in this case, we eliminate the dependence on $\gamma$, but the trade-off is a worse dependence on $T$.
We obtain the $O\left( \sqrt{n^3 T} \right)$ by analyzing the Two-Phase Algorithm with the choice of arms $\Sigma=\left\{ e_{1},\ldots,e_{n}\right\}$, the standard basis, and the scheduling function $g(l) = \lfloor \dfrac{l}{n} \rfloor$.

The proof can be decomposed into several parts. 

\begin{itemize}
	\item First, we have already shown that the probability of an epoch being bad decreases exponentially in the epoch number, in Lemma \ref{lemma bad epoch}.
	\item Then, for a good epoch, the probability that $\hat{\alpha}$ deviates from the true value $\alpha^*$ also decreases exponentially in the epoch number. 
	\item Large deviations of this estimate can be related to large errors in the central angle between the estimated value $\hat{z}_l$ and the true value $z^*$.
	\item Large regret implies large deviations in this central angle.
	\item Finally, the total expected regret can be bounded using the fact $E[r_{2,l}] = \int_0^1 P(r_{2,l} > \delta) d \delta$.
\end{itemize}

Define $\Theta_l$ to be the central angle between $\hat{z}_l$ and $z^*$ when $\hat{z}_l \ne \boldsymbol{0}_n$, and $\pi$ otherwise, i.e.
$$\Theta_l=\begin{cases}
\arccos \left( \dfrac{\left( \hat{z}_l \right) ^T z^*}{\left\Vert \hat{z}_l \right\Vert \cdot \left\Vert z^* \right\Vert} \right),
 & \hat{z}_l \ne \boldsymbol{0}_n \\
\pi, & \hat{z}_l = \boldsymbol{0}_n
\end{cases}$$

Recall that $G_l$ denotes the event that epoch $l$ is a good epoch.

\begin{lem}\label{lemma infinite regret deviation implies angle deviation}
$$ r_{2,l} > \delta \implies \Theta_l > \sqrt{\dfrac{8 \delta}{\left\Vert z^{*}\right\Vert } } .$$
\end{lem}
Proof:
\begin{align*}
 & r_{2,l}>\delta \\
\iff & f\left(\left(u^{*}\right)^{T}z^{*}\right)-f\left( \left( \hat{u}_l \right) ^{T}z^{*} \right)>\delta\\
\iff & f\left(\dfrac{\left(z^{*}\right)^{T}z^{*}}{\left\Vert z^{*}\right\Vert }\right)-f \left( \left( \hat{u}_l \right) ^{T}z^{*} \right)>\delta \\
\iff & f\left(\left\Vert z^{*}\right\Vert \right)-f\left(\cos\left(\Theta_l\right)\left\Vert z^{*}\right\Vert \right)>\delta \tag{1} \\
\implies & \left\Vert z^{*}\right\Vert \left(1-\cos\left(\Theta_l\right)\right) > \dfrac{\delta}{\min_x f'\left( x \right)} = 4 \delta \\
\implies & \dfrac{\Theta_l^{2}}{2}>\dfrac{4\delta}{\left\Vert z^{*}\right\Vert } \tag{2} \\
\iff & \Theta_l> \sqrt{\dfrac{8 \delta}{\left\Vert z^{*}\right\Vert } } ,
\end{align*}
where (1) is since $\cos(\Theta_l) = \dfrac{\left( \hat{z}_l \right) ^T z^*}{\left\Vert \hat{z}_l \right\Vert \cdot \left\Vert z^* \right\Vert}$, and (2) is since $\cos(x) \ge 1-\dfrac{x^2}{2},\ \forall x$.
\begin{flushright}
$\square$
\end{flushright}

\begin{figure}
\begin{center}
\includegraphics{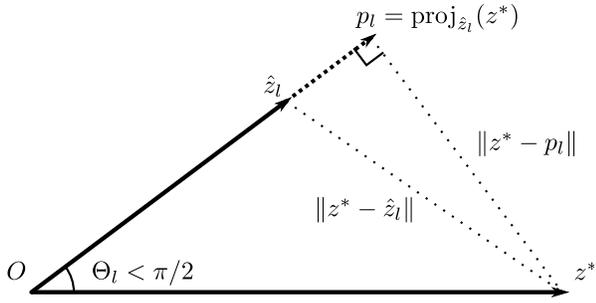}
\end{center}
\caption{Schematic diagram illustrating the relationships between 
$\left\Vert \hat{z}_{l} - z^{*}\right\Vert$, $\left\Vert z^* - p_l \right\Vert$, and $\Theta_l$.
}
\label{fig:infinite zhat zstar projection}
\end{figure}

\begin{lem}\label{lemma infinite angle deviation implies zhat deviation}
With the choice $\Sigma=\left\{ e_{1},\ldots,e_{n}\right\}$,
$$
\Theta_l > \theta \implies \exists u \in \Sigma \mbox{ s.t.}
\left| u^T \hat{z}_{l} - u^T z^{*}\right| \ge \dfrac{\theta}{\pi\sqrt{n}}\cdot\left\Vert z^{*}\right\Vert.
$$
\end{lem}
Proof: 
Suppose that $\Theta_l > \theta$. There are two cases: $$\Theta_l \in [0, \pi / 2] \mbox{, or } \Theta_l \in (\pi/2, \pi] .$$
In the first case, define $p$ to be the vector projection of $z^*$ onto  $\hat{z}_l$; see Figure \ref{fig:infinite zhat zstar projection}.
Then, 
\begin{align*}
\left\Vert \hat{z}_{l} - z^{*}\right\Vert & \ge \left\Vert z^* - p \right\Vert \\
 & = \sin(\Theta_l)\cdot\left\Vert z^{*}\right\Vert \\
 & > \sin(\theta)\cdot\left\Vert z^{*}\right\Vert.
\end{align*}
In the second case, when $\Theta_l > \pi / 2$,
$$\left\Vert \hat{z}_{l} - z^{*}\right\Vert \ge \left\Vert z^{*}\right\Vert.$$
Note that this is true even in the special case of $\hat{z}_l=\boldsymbol{0}_n$, by our definition of $\Theta_l$.
Then, $\forall \theta \in [0, 2\pi]$, 
\begin{align*}
& \Theta_l > \theta \\
\implies & \left\Vert \hat{z}_{l} - z^{*}\right\Vert \ge \dfrac{\theta}{\pi}\cdot\left\Vert z^{*}\right\Vert \\
\implies & \exists i \in \{1, 2, \ldots , n\} \mbox{ s.t.} \\
& \quad \left| \left( e_i \right)^T \hat{z}_{l} - \left( e_i \right)^T z^{*} \right| \ge \dfrac{\theta}{\pi\sqrt{n}}\cdot\left\Vert z^{*}\right\Vert \\
\implies & \exists u \in \Sigma \mbox{ s.t.} \\
& \quad \left| u^T \hat{z}_{l} - u^T z^{*}\right| \ge \dfrac{\theta}{\pi\sqrt{n}}\cdot\left\Vert z^{*}\right\Vert.
\end{align*}
\begin{flushright}
$\square$
\end{flushright}

\begin{lem}\label{lemma infinite zhat deviation implies alpha deviation}
In a good epoch,
\begin{align*}
& \exists \tilde{u} \in \Sigma = \{e_1, e_2, \ldots, e_n\}, \delta_2 \in (0, 1], \mbox{ s.t. } \\
& \left| \tilde{u}^T \hat{z}_{l} - \tilde{u}^T z^{*}\right| \ge \delta_2 \cdot \left\Vert z^{*}\right\Vert \\
\implies & \left| \hat{\alpha}_{\tilde{u},l} -\alpha_{\tilde{u}}^* \right| \ge \delta_2 \cdot\left\Vert z^{*}\right\Vert \cdot f'\left(2\left\Vert z^{*}\right\Vert \right) .
\end{align*}
\end{lem}
Proof: 
Suppose the current epoch is good, and
\begin{align*}
& \exists \tilde{u} \in \Sigma = \{e_1, e_2, \ldots, e_n\}, \delta_2 \in (0, 1], \mbox{ s.t. } \\
& \left| \tilde{u}^T \hat{z}_{l} - \tilde{u}^T z^{*}\right| \ge \delta_2 \cdot \left\Vert z^{*}\right\Vert .
\end{align*}

Since the current epoch is good, $f^{-1}\left(\hat{\alpha}_{u,l}\right)$ exists $\forall u \in \Sigma$.
By construction, $\Sigma$ is full rank, and
$$ 
\hat{z}_l = \left(\Sigma^{T}\right)^{-1}\left[\begin{array}{c}
														f^{-1}\left(\hat{\alpha}_{\Sigma_1,l}\right)\\
														\vdots\\
														f^{-1}\left(\hat{\alpha}_{\Sigma_n,l}\right)\end{array}\right].
$$
Thus, $\forall u \in \Sigma$, 
\begin{align*}
& \hat{\alpha}_{u,l} = f \left( u^T \hat{z}_l \right) \mbox{ and}\\
& \alpha_{u}^* = f \left( u^T z^{*} \right).
\end{align*}

Let $\tilde{\beta} = \tilde{u}^T z^* + \delta_2 \cdot \left\Vert z^{*}\right\Vert \cdot \mbox{sgn} \left( \tilde{u}^T \hat{z}_l - \tilde{u}^T z^{*} \right)$. Then,

\begin{align*}
\left|\hat{\alpha}_{\tilde{u},l} -\alpha_{\tilde{u}}^*\right|
= & \left| f \left( \tilde{u}^T \hat{z}_l \right) - f \left( \tilde{u}^T z^{*} \right) \right| \\
\ge & \left| f \left( \tilde{\beta} \right) - f \left( \tilde{u}^T z^{*} \right) \right| \tag{1} \\
\ge & \left| \tilde{\beta} - \tilde{u}^T z^{*} \right| \cdot \min_{x \in \left[ \tilde{\beta}, \tilde{u}^T z^{*} \right]} f'(x) \tag{2} \\
\ge & \delta_2 \cdot \left\Vert z^{*}\right\Vert \cdot \min_{x \in \left[ -2 \left\Vert z^{*}\right\Vert, 2 \left\Vert z^{*}\right\Vert \right] } f'(x) \tag{3} \\
= & \delta_2 \cdot\left\Vert z^{*}\right\Vert \cdot f'\left(2\left\Vert z^{*}\right\Vert \right), \tag{4}
\end{align*}
where (1) is since $f$ is strictly increasing and $$\tilde{\beta} \in \left[ \tilde{u}^T z^*, \tilde{u}^T \hat{z}_l \right] ,$$
(2) is since $f'$ is continuous, 
(3) is since $\delta_2 \le 1$ and
$$\left\Vert u^T z^* \right\Vert \le \left\Vert z^{*}\right\Vert \ \forall u \in \Sigma ,$$
and thus $\left[ \tilde{\beta}, \tilde{u}^T z^{*} \right] \subset  \left[ -2 \left\Vert z^{*}\right\Vert, 2 \left\Vert z^{*}\right\Vert \right]$,
and (4) is since $f'$ is unimodal and symmetric.
\begin{flushright}
$\square$
\end{flushright}

\begin{lem}\label{lemma infinite alpha deviation probability}
\begin{align*}
& P \left( \exists u \in \Sigma : \left|\hat{\alpha}_{u,l} -\alpha_u^*\right| \ge \delta_3 \right) \\
\le & 2n \cdot \exp \left( -2l \cdot \left( \delta_3 \right)^2 \right).
\end{align*}
\end{lem}
Proof: 
Note that  $ E[\hat{\alpha}_{u,l}] = \alpha_u^*, \ \forall l \in \mathbb{N}_1, u \in \Sigma$. Then, for any $u \in \Sigma$, by the Hoeffding bound,
$$P \left(\left|\hat{\alpha}_{u,l} -\alpha_u^*\right| \ge \delta_3 \right) \le  2 \cdot \exp \left( -2l \cdot \left( \delta_3 \right)^2 \right) .$$
Applying a union bound over all $u \in \Sigma$ yields the desired result.
\begin{flushright}
$\square$
\end{flushright}

\begin{lem}\label{lemma infinite per timestep regret phase 2}
\begin{align*}
E\left[r_{2,l} | G_l \right] \le \dfrac{\pi^2 n^2}{8 l\cdot \left\Vert z^{*}\right\Vert \cdot f'\left(2\left\Vert z^{*}\right\Vert \right)^{2}}.
\end{align*}
\end{lem}
Proof:
Combining Lemmas \ref{lemma infinite regret deviation implies angle deviation}, \ref{lemma infinite angle deviation implies zhat deviation}, \ref{lemma infinite zhat deviation implies alpha deviation} and \ref{lemma infinite alpha deviation probability}, we have that
\begin{align*}
& P \left( r_{2,l} > \delta \right | G_l) \\
\le & P \left( \Theta_l > \sqrt{\dfrac{8 \delta}{\left\Vert z^{*}\right\Vert } } | G_l \right) \\
\le & P \left( \exists u \in \Sigma :
\left| u^T \hat{z}_{l} - u^T z^{*}\right| \ge \sqrt{\dfrac{8 \delta}{\pi^2 n \left\Vert z^{*}\right\Vert } } \cdot\left\Vert z^{*}\right\Vert | G_l \right) \\
\le & P \left( \exists u \in \Sigma : \left| \hat{\alpha}_{u,l} -\alpha_{u}^* \right| \ge \sqrt{\dfrac{8 \delta \cdot \left\Vert z^{*}\right\Vert }{\pi^2 n  } } \cdot f'\left(2\left\Vert z^{*}\right\Vert \right) \right) \\
\le & 2n \cdot \exp \left( -l \cdot \dfrac{16 \delta}{\pi^2 n } \cdot \left\Vert z^{*}\right\Vert \cdot f'\left(2\left\Vert z^{*}\right\Vert \right)^2 \right) .
\end{align*}
Let $k_3$ denote the constants independent of $l$ and $n$, namely 
$$k_3 = \dfrac{16}{\pi^2} \cdot \left\Vert z^{*}\right\Vert \cdot f'\left(2\left\Vert z^{*}\right\Vert \right)^{2}.$$
Then,
\begin{align*}
E\left[r_{2,l} | G_l \right] & = \int_{0}^{1}P \left( r_{2,l}>\delta | G_l \right) d \delta \\
 & \le \int_{0}^{1} 2n\cdot\exp\left(-\delta \cdot\dfrac{ k_3 l }{ n } \right)d \delta \\ 
 & = 2n\cdot\dfrac{- n}{ k_3 l}\cdot\left[\exp\left(-\delta \cdot\dfrac{ k_3 l }{ n } \right)\right]_{0}^{1} \\ 
 & \le \dfrac{2 n^2}{ k_3 l }.
\end{align*}

\begin{flushright}
$\square$
\end{flushright}

\begin{thm}\label{theorem infinite regret}
For the Two-Phase Algorithm on a unit sphere of arms in $\mathbb{R}^n$, we have the following bound on the total expected regret up to time-horizon $T$:
$$
E\left[R_{T}\right] \le \left( 1 + \dfrac{2}{ k_3 } \right) \cdot nL + 2 \sum_{l=1}^{L} l \cdot \exp \left( -k_4 l \right) .
$$

\end{thm}
Proof:
Since the final epoch may be only partially finished, we will lower-bound the total time with the number of timesteps in all prior epochs. Decomposing into Phase 1 and Phase 2, we have
\begin{align*}
T & \ge \sum_{l=1}^{L-1} \left\{ n + \lfloor \dfrac{l}{n} \rfloor \right\} \\ 
& \ge \sum_{l=1}^{L-1} \dfrac{l}{n} \\
& \ge \dfrac{(L-1)^2}{2n}.
\end{align*}
Equivalently,
$$L \le \sqrt{2nT} + 1.$$
By Lemmas \ref{lemma bad epoch} and \ref{lemma infinite per timestep regret phase 2}, we can upper-bound the expected total Phase 2 regret,
\begin{align*}
& E[R_{2,T}] \\
\le & \sum_{l=1}^{L} \left\{ \Big[ P(G_l) \cdot E[r_{2,l}|G_l] + P( \neg G_l) \cdot E[r_{2,l}| \neg G_l] \Big] \cdot \lfloor \dfrac{l}{n} \rfloor \right\} \\
\le & \sum_{l=1}^{L} \left\{ \left[ P(G_l) \cdot \dfrac{2 n^2}{ k_3 l } + P( \neg G_l) \cdot 1 \right] \cdot \dfrac{l}{n} \right\} \\
\le & \sum_{l=1}^{L} \left\{ \left[ \dfrac{2 n^2}{ k_3 l } + 2n \cdot \exp\left( -k_1 l \right) \right] \cdot \dfrac{l}{n} \right\} \\
\le & \dfrac{2nL}{ k_3 } + 2 \sum_{l=1}^{\infty} l \cdot \exp \left( -k_1 l \right) .
\end{align*}
Then, using Lemma \ref{lemma total phase 1},
\begin{align*}
E[R_{T}] = & E[R_{1,T}] + E[R_{2,T}] \\
\le & \left( \alpha_V^* + \dfrac{2}{ k_3 } \right) \cdot nL + 2 \sum_{l=1}^{\infty} l \cdot \exp \left( -k_1 l \right) \\
\le & \left( \alpha_V^* + \dfrac{2}{ k_3 } \right) \cdot \left( \sqrt{2n^3 T} + n \right) \\
& + 2 \sum_{l=1}^{\infty} l \cdot \exp \left( -k_1 l \right) .
\end{align*}
\begin{flushright}
$\square$
\end{flushright}

\begin{cor}
For the Two-Phase Algorithm on a unit sphere of arms in $\mathbb{R}^n$, we have the following asymptotic bound on the expected total regret up to timestep $T$:
$$E \left[ R_{T} \right] \in O\left( \sqrt{n^3 T} \right).$$
\end{cor}
Proof:
By Theorem \ref{theorem infinite regret}, as a function of $T$,
\begin{align*}
E[R_T] = & \left( \alpha_V^* + \dfrac{2}{ k_3 } \right) \cdot \left( \sqrt{2n^3 T} + n \right) \\
& + 2 \sum_{l=1}^{\infty} l \cdot \exp \left( -k_1 l \right) \\
\in \  & O \left( \sqrt{n^3 T} \right),
\end{align*}
since $\alpha_V^* \le 1$, and $k_1 , k_3 > 0 $ are both constants dependent only upon $\left\Vert z^{*}\right\Vert$.
\begin{flushright}
$\square$
\end{flushright}

\begin{remark}
The choice of scheduling function $g$ is not restricted to be $\left\lfloor \frac{l}{n} \right\rfloor$; the dependence on $n$ can be altered to change the trade-off between the constants in front of $\sqrt{T}$ and $ \sum_{l=1}^{\infty} l \cdot \exp \left( -k_1 l \right)$. 
That is, the asymptotics can be improved at the expense of short time-horizon performance. Furthermore, if the time-horizon is known in advance, then the scheduling function can be chosen to minimize the sum of these two terms, just as in the finite arm case.
\end{remark}


\addtolength{\textheight}{-47mm}   

\section{CONCLUSIONS}
We have proposed a class of parametrized multi-armed bandit problems,
in which the reward distribution is Bernoulli and independent across
arms and across time, with a parameter that is a non-linear function
of the scalar quality of an arm. The real-valued qualities are 
inner products between the unknown preference and known attribute vectors. Under
this model, we are able to capture the fundamentally binary choice
inherent in certain online machine learning problems. 

Our proposed algorithm achieves an asymptotic expected 
total regret of $O \left( n \cdot g^{-1}(T) \right)$ for any function 
$\left. g^{-1}(T) \in \omega \left( \log(T) \right) \right.$ in the finite arm case,
and $O \left( \sqrt{n^3T} \right)$ in the infinite arm, unit circle case.
This is in contrast to the $\Omega \left( m \log(T) \right)$ lower-bound of Lai and Robbins, and the 
$\Omega \left(T^{\frac{n+1}{n+2}} \right)$ lower-bound of Kleinberg \textsl{et al.}
In both cases, the additional assumption of structure (linearly correlated instead of independent, and logistic function of linear instead of Lipschitz, respectively) can be used to out-perform optimal algorithms which do not account for this structure.
We conjecture that the lower-bounds on our problem are 
$\left. \Omega(n \cdot \log(T)) \right.$ and $\Omega(\sqrt(T))$ for the finite and infinite arm cases, respectively; if true, then this simple algorithm's performance is nearly optimal.

Finally, our algorithm can be implemented very efficiently, since the storage requirements are $O(n)$ and thus do not scale with either the number of arms or the time-horizon. Also, since the exporation and expoitation phases are decoupled, the only history-dependent part of the algorithm, the optimization to determine which arm to pull, is only performed during a small number of timesteps (approximately $O(\log(T))$ and $O(\sqrt{T})$ for finite and infinite arm cases, respectively). In the infinite arm, unit circle case, this optimization itself is simply the normalization of the current estimate $\hat{z}_l$.

The basic idea of increasing the length of epochs is similar to that of UCB2, but because our algorithm uses a global count of the epoch instead of local counts for each arm, it is applicable to infinite arm problems.
Finally, we note that several extensions to this work are possible; multiple plays and time-dependent $U$ and $z^{*}$ would  be directly applicable for e-commerce applications.



\bibliographystyle{IEEEtran}
\bibliography{IEEEabrv,paper}
\end{document}